\documentclass[a4paper,conference]{IEEEtran}
\usepackage{packages}
\begin{document}
\title{The Multiscale Bowler-Hat Transform for Vessel Enhancement in 3D Biomedical Images}
\author{\IEEEauthorblockN{\c{C}i\u{g}dem~Sazak}
	\IEEEauthorblockA{Department of Computer Science \\ Durham University, UK\\
		Email: cigdem.sazak@durham.ac.uk }
	\and
	\IEEEauthorblockN{Carl~J.~Nelson}
	\IEEEauthorblockA{School of Physics and Astronomy \\Glasgow University, UK\\
		Email: chas.nelson@glasgow.ac.uk}
	\and
	\IEEEauthorblockN{Boguslaw~Obara$^*$}
	\IEEEauthorblockA{Department of Computer Science \\ Durham University, UK\\
	Email: boguslaw.obara@durham.ac.uk }}

\maketitle
\begin{abstract}
 Enhancement and detection of 3D vessel-like structures has long been an open problem as most existing image processing methods fail in many aspects, including a lack of uniform enhancement between vessels of different radii and a lack of enhancement at the  junctions. 
 Here, we  propose a method based on mathematical morphology to enhance 3D vessel-like structures in biomedical images. The proposed method, 3D bowler-hat transform, combines sphere and line structuring elements to enhance vessel-like structures. The proposed method is validated on synthetic and real data, and compared with state-of-the-art methods. 
 Our results show that the proposed method achieves a high-quality vessel-like structures enhancement in both synthetic and real biomedical images, and is able to cope with variations in vessels thickness throughout vascular networks while remaining robust at junctions.
\end{abstract}

\section{Introduction}\label{sec:intro}

Automatic detection of vessel-like structures is one of the fundamental procedures in many 3D biomedical image processing applications, where they are used to understand important vascular networks, such as cytoskeletal networks, blood vessels, airways, and other similar fibrous tissues. 
Reliable detection and then accurate analysis of these vascular networks strongly relies on robust vessel-like structures enhancement methods. Several such methods have been proposed and investigated for various types of biomedical images such as: 
blood vessels~\cite{Benmansour2011,frangi1998multiscale}, neurons~\cite{al2008improved}, microtubules~\cite{hadjidemetriou2005segmentation} and others~\cite{bijith2017image,ortiz2012ultrasound}.
Nevertheless, most of the vessel-like structures enhancement methods still suffer from unresolved problems such as losing signals at the junctions or false vessel effects~\cite{sazak2017contrast}.

In this paper, we extended method 2D to 3D vessel-like structures enhancement method, called the 3D bowler-hat transform. The proposed method is based on a recently developed 2D image filtering method exploring a concept of mathematical morphology~\cite{sazak2017multiscale}. 
We qualitatively and quantitatively validate and compare the proposed method with the state-of-the-art methods using a range of synthetic and real biomedical images.
Our results show that the proposed method produces a high-quality vessel enhancement, especially at junctions in both synthetic and real images. The method is suitable to be applied to a variety of biomedical image types without requiring prior preparation or tuning. Finally, we make our method available online, along with source code and all test functions. 

\section{Related Work} \label{sec:related}\vspace{-5pt}
A considerable amount of work has been conducted on the enhancement and segmentation of vessel-like structures~\cite{zhang2014blood,kerkeni2016coronary}. In this section we summarise them into three categories: Hessian-based enhancement, phase-congruency-based enhancement, and morphological enhancement, and the representative state-of-the-art works are reviewed under each categories.

\subsection{Hessian-based Enhancement Methods}\label{subsec:hessian}
Frangi et al.~\cite{frangi1998multiscale} proposed 2/3D vessel-like structures enhancement in biomedical images by exploring the relationships between eigenvectors and eigenvalues of a Hessian matrix. The Hessian is constructed with responses of a set of matching filters, defined by second-order derivatives of the Gaussian function, convolved with the image. Three most common measurements proposed to date are vesselness, neuriteness and regularized volume ratio. 

\subsubsection{Vesselness}\label{subsec:ves} 
Vesselness measure~\cite{frangi1998multiscale} is a function of the eigenvalues of the Hessian matrix of the image data. The eigenvalues of the Hessian matrix correspond to the second derivatives of the image data in the direction of the associated eigenvector. In general, vesselness fails at vessels junctions due to the low filters responses. 

\subsubsection{Neuriteness}\label{subsec:neurite}
As an alternative to vesselness, neuriteness measure modifies the Hessian matrix by adding a new parameter to improve vessel-like structures enhancement in 2D biomedical images~\cite{meijering2004design}. This work was then extended for use in 3D biomedical images by~\cite{al2008improved}. 
Neuriteness, in the same way as the vesselness, fails at vessels junctions due to the low filters responses.

\subsubsection{Regularized Volume Ratio}\label{subsec:volrat}
A problem with Hessian-based methods such as vesselness or neuriteness is the direct proportionality of the output to the eigenvalues. Due to eigenvalue heterogeneity within objects and variation in eigenvalue magnitude, this proportionality results in non-uniform enhancement.
In~\cite{JPLetal2016}, authors attempt to solve this problem by deriving a modification of the volume ratio with a regularised eigenvalue to ensure robustness to small changes in magnitude.

\subsection{Phase Congruency-based Enhancement Methods}\label{subsec:pc}
Most image enhancement methods have a common problem of image contrast and spatial magnification dependency, which causes low contrast vessels to be missed~\cite{kovesi1996invariant,kovesi1999image}. 
To overcome this problem,~\cite{morrone1986mach} proposes a contrast-independent image features enhancement method exploring a concept called a phase congruency. The phase congruency compares the weighted alignment of the Fourier components of the image with the sum of the Fourier components~\cite{Wang2008301,ferrari2011detection}.
 
In similar way as with the Hessian matrix concept, a Phase Congruency Tensor (PCT) is proposed to represent local structures in the image; first in 2D by~\cite{obara2012contrast} and then in 3D by~\cite{sazak2017contrast}. 
Then, eigenvalues and eigenvectors of the PCT are calculated and used to define PCT vesselness and PCT neuriteness. A major drawback of phase-based methods is the complexity of the parameter space to calculate the PCT. 

\subsection{Enhancement with Mathematical Morphology}\label{subsec:morph}
Another class of image enhancement methods is based on mathematical morphology, which has been used for several challenges~\cite{CWRTetal2002,JWR2004,Su2014,morales2013automatic}. 
To enhance vessel-like structures, Zana and Klein~\cite{ZK2001} proposes a use of morphological transforms. This method assumes that vessels are linear, connected and have smooth variations of curvature along the peak of the structure. First, a sum of top-hats is computed using linear structuring elements at different angles; then a curvature measure is calculated using a Laplacian of Gaussian, and finally, both of them are combined to reduce noise and enhance vessel-like structures in an image.

\subsection{Limitations and Challenges}
Many vessel-like structures enhancement methods still fail in ways that compromise their use in automated detection and analysis pipelines. For example, contrast variations cause low-accuracy enhancement, and high noise levels leads to poor enhancement and the 'false vessel' effect. Another common issue is dealing with junctions, where most Hessian- or PCT-based methods suppress the 'disk-like' features at a junction leading to a loss of extracted network connectivity. Further, some methods are computationally expensive or have a complex set of parameters that can be time-consuming to manually fine-tune.
\vskip-15pt
\section{Method}\label{sec:proposed:method}
In this section, we introduce a 3D extension of a recently introduced mathematical morphology-based 2D method for vessel-like structures enhancement called the bowler-hat transform~\cite{sazak2017multiscale}.
While explaining the details of the proposed method, we point out the concepts that allow us to address the major drawbacks of existing, state-of-the-art vessel-like structures enhancement methods.
 
\subsection{Mathematical Morphology}
Mathematical morphology has been extensively used in image processing and image analysis~\cite{serra1986introduction,soille2013morphological}.
Mathematical morphology uses structuring elements and concepts from set theory to describe features of interest in images. Most morphological operations are based on two basic operations: dilation and erosion. These two operations take the image as an input and dilate or erode components within the particular area with structuring element.
While erosion extends dark areas decreasing bright areas, dilation expands bright areas and decrease dark areas.\\
Using these two operators, two further operations can be defined, that are called opening and closing.
The closing preserves bright structures while suppressing dark patterns meantime the opening maintains dark structures and patterns and suppressing bright features. 

\input{./tex/figure-flow}

\subsection{Proposed Method}

The proposed method is explained and summarized with a small example in~\Cref{fig:pro:method}.  
The 3D bowler-hat transform combines two banks of different structuring elements: a bank of spherical structuring elements with varying diameter~(\Cref{fig:shp4}) and a bank of orientated line structuring elements with varying length and directions~(\Cref{fig:orientation}).

First, we create a bank of morphological openings of a 3D input image $I$ with spherical structuring elements $S_{sphere}^{d}$ of diameter $d \in {[1,d_{max}]}$, where $d_{max}$ is expected maximum size of vessel-like structures in a given image $I$. After every morphological opening of the image $I$, vessel-like structures smaller than $d$ are eliminated and the ones larger than $d$ remain.

As a result, a 4D image stack, for all $d \in {[1,d_{max}]}$, is constructed as:

\begin{equation}
\{I_{sphere}\} = \{I \circ S_{sphere}^{d}\}, \quad \forall r \in [1,d_{max}].
\end{equation}

Then, another bank of morphological openings of the input image $I$ is performed with line structuring elements $S_{line}^{d,\mathbf{v}}$ of lengths $d$, $\forall d \in [1,d_{max}]$, and of directions defined as follows:

\begin{equation}
\mathbf{v} = (\theta_k,\phi_k), \quad  \forall k \in [1,N].
\end{equation}

Direction $\left( \theta_k,\phi_k \right)$ is defined as a $k^{th}$ point from $N$ uniformly distributed points on the unit sphere, and more details can be found in~\cite{koay2011analytically}.
After every morphological opening of the image $I$ with a line structuring element $S_{line}^{d,\mathbf{v}}$, vessel-like structures smaller than $d$ along direction $\mathbf{v}$ are eliminated but all vessel-like structures that are longer than $d$ along direction $\mathbf{v}$ remain. This step results in a 5D image stack for all lengths $d$ and all directions $\mathbf{v}$:

\begin{equation}
\{\{I_{line}\}\} = \{\{I \circ S_{line}^{d,\mathbf{v}}\}\}, \forall d \in [1,d_{max}],  \forall k \in [1,N].
\end{equation}

Then, for each length $d$, a pixel-wise maximum across all directions $\mathbf{v}$ is calculated resulting in a 4D image stack:

\begin{equation}
\{I_{line}\} = \{\max_{k \in [1,N]}|\{I \circ S_{line}^{d,\mathbf{v}}\}\arrowvert\}, \quad \forall r \in [1,d_{max}].
\end{equation}

The enhanced image is then produced by taking maximum stack-wise difference at each pixel,

\begin{equation}
I_{enhanced} = \max_{r \in [1,d_{max}]}|\{I_{sphere} - I_{line}\}\arrowvert.
\end{equation}

With the 3D bowler-hat transform, areas that are dark (background) in the original image remain dark due to the use of openings; blob-like bright objects (undesired foreground features) are suppressed as the sphere-based and line-based opening gives similar values; and tube-like bright objects (desired foreground features) are enhanced due to the large difference between sphere-based and longer line-based openings. 
To assign an appropriate $d_{max}$, expected maximum vessel-like structures size in the image, allows the identification of most of the vessel-like structures and junctions, something that many other vessel enhancement methods fail to do. This is due to the ability to fit longer line-based structural elements within the junction area. In~\Cref{sec:results} we illustrate the key advantages of the proposed method over other vessel-like structures enhancement methods.
\section{Results and Discussions} \label{sec:results}

In this section, we qualitatively and quantitatively validate the efficiency of the proposed method using a range of synthetic and real biomedical image datasets. We then compare the proposed method with the state-of-the-art vessel-like structures enhancement methods such as Hessian-based vesselness~\cite{frangi1998multiscale}, neuriteness~\cite{lindeberg2013scale} and volume ratio~\cite{JPLetal2016}, and PCT-based vesselness \& neuriteness~\cite{sazak2017contrast}.

\subsection{Quantitative Validation}
While a visual examination can give some subjective information regarding the effectiveness of the vessel enhancement method, a form a quantitative validation is also required. To compare the proposed method with the other state-of-the-art algorithms, we have chosen to calculate the Receiver Operating Characteristic (ROC) curve and the Area Under the Curve (AUC) and more details can be found in~\cite{hajian2013receiver}. 

\newcommand\sfigure{0.23}
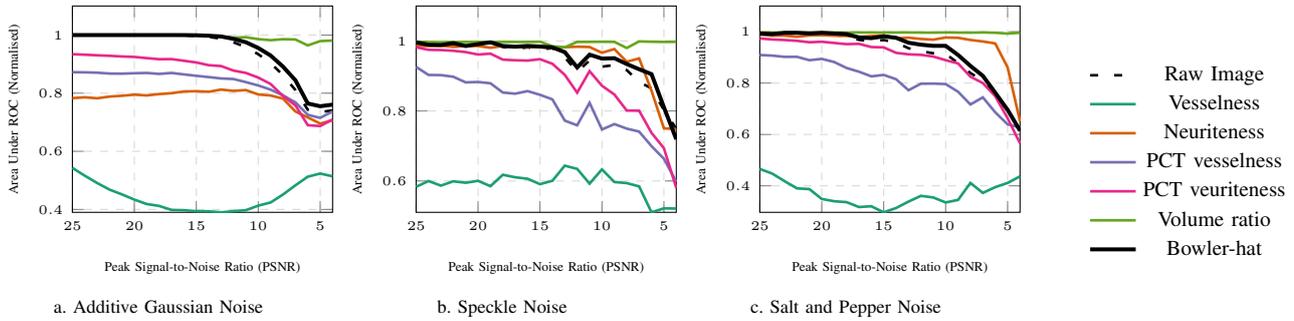
\begin{figure*}
	\centering
	\begin{subfigure}{\sfigure\linewidth}
		\begin{tikzpicture}
		\begin{axis}[
		width =1.2\linewidth,
		xlabel={Peak Signal-to-Noise Ratio (PSNR)},
		ylabel={Area Under ROC (Normalised)},
		ylabel near ticks,
		label style={font=\tiny},
		tick label style={font=\tiny},
		ytickmin=0, ymax=1.1,
		xtickmin=0, xtickmax=25, x dir=reverse,
		enlargelimits=false,
		legend entries = {Raw Image, Vesselness, Neuriteness, PCT vesselness, PCT veuriteness, Volume ratio, Bowler-hat},
		legend columns = 1,
		legend style={font=\footnotesize, draw=none,},
		legend to name=leg2,
		grid=major, 
		grid style={dashed,gray!30}, 
		]
		\addplot[color=black,loosely dashed,line width=1pt] table [x index=0, y index=1, col sep=comma] {image/dat-files/gaussian.dat};
		\addplot[color=brewerDark1,line width=1pt] table [x index=0, y index=3, col sep=comma] {image/dat-files/gaussian.dat};
		\addplot[color=brewerDark2,line width=1pt] table [x index=0, y index=4, col sep=comma] {image/dat-files/gaussian.dat};
		\addplot[color=brewerDark3,line width=1pt] table [x index=0, y index=5, col sep=comma] {image/dat-files/gaussian.dat};
		\addplot[color=brewerDark4,line width=1pt] table [x index=0, y index=6, col sep=comma] {image/dat-files/gaussian.dat};
		\addplot[color=brewerDark5,line width=1pt] table [x index=0, y index=7, col sep=comma] {image/dat-files/gaussian.dat};
		\addplot[color=black,line width=1.5pt] table [x index=0, y index=2, col sep=comma] {image/dat-files/gaussian.dat};
		\end{axis}
		\end{tikzpicture}
		\caption{Additive Gaussian Noise}\label{fig:psnr:gauss}
	\end{subfigure}\quad
	\begin{subfigure}{\sfigure\linewidth}
		\begin{tikzpicture}
		\begin{axis}[
		width =1.2\linewidth,
		xlabel={Peak Signal-to-Noise Ratio (PSNR)},
		ylabel={Area Under ROC (Normalised)},
		ylabel near ticks,
		label style={font=\tiny},
		tick label style={font=\tiny},
		ytickmin=0, ymax=1.1,
		xtickmin=0, xtickmax=25, x dir=reverse,
		enlargelimits=false,
		grid=major, 
		grid style={dashed,gray!30}, 
		]
		\addplot[color=black,loosely dashed,line width=1pt] table [x index=0, y index=1, col sep=comma] {image/dat-files/speckle.dat};
		\addplot[color=brewerDark1,line width=1pt] table [x index=0, y index=3, col sep=comma] {image/dat-files/speckle.dat};
		\addplot[color=brewerDark2,line width=1pt] table [x index=0, y index=4, col sep=comma] {image/dat-files/speckle.dat};
		\addplot[color=brewerDark3,line width=1pt] table [x index=0, y index=5, col sep=comma] {image/dat-files/speckle.dat};
		\addplot[color=brewerDark4,line width=1pt] table [x index=0, y index=6, col sep=comma] {image/dat-files/speckle.dat};
		\addplot[color=brewerDark5,line width=1pt] table [x index=0, y index=7, col sep=comma] {image/dat-files/speckle.dat};
		\addplot[color=black,line width=1.5pt] table [x index=0, y index=2, col sep=comma] {image/dat-files/speckle.dat};
		\end{axis}
		\end{tikzpicture}
		\caption{Speckle Noise}\label{fig:psnr:speck}
	\end{subfigure}\quad
	\begin{subfigure}{\sfigure\linewidth}
		\begin{tikzpicture}
		\begin{axis}[
		width =1.2\linewidth,
		xlabel={Peak Signal-to-Noise Ratio (PSNR)},
		ylabel={Area Under ROC (Normalised)},
		ylabel near ticks,
		label style={font=\tiny},
		tick label style={font=\tiny},
		ytickmin=0, ymax=1.1,
		xtickmin=0, xtickmax=25, x dir=reverse,
		enlargelimits=false,
		grid=major, 
		grid style={dashed,gray!30}, 
		]
		\addplot[color=black,loosely dashed,line width=1pt] table [x index=0, y index=1, col sep=comma] {image/dat-files/pepper.dat};
		\addplot[color=brewerDark1,line width=1pt] table [x index=0, y index=3, col sep=comma] {image/dat-files/pepper.dat};
		\addplot[color=brewerDark2,line width=1pt] table [x index=0, y index=4, col sep=comma] {image/dat-files/pepper.dat};
		\addplot[color=brewerDark3,line width=1pt] table [x index=0, y index=5, col sep=comma] {image/dat-files/pepper.dat};
		\addplot[color=brewerDark4,line width=1pt] table [x index=0, y index=6, col sep=comma] {image/dat-files/pepper.dat};
		\addplot[color=brewerDark5,line width=1pt] table [x index=0, y index=7, col sep=comma] {image/dat-files/pepper.dat};
		\addplot[color=black,line width=1.5pt] table [x index=0, y index=2, col sep=comma] {image/dat-files/pepper.dat};
		\end{axis}
		\end{tikzpicture}
		\caption{Salt and Pepper Noise}\label{fig:psnr:pepp}
	\end{subfigure}\quad
	\begin{subfigure}{\sfigure\linewidth}
		\centering
		\pgfplotslegendfromname{leg2}
	\end{subfigure}
	\caption{Mean AUC for the input image and the image enhanced by the proposed method and by the state-of-the-art methods with different peak signal-to-noise ratios (PSNRs) for three different noise types: (\subref{fig:psnr:gauss}) additive Gaussian noise, (\subref{fig:psnr:speck}) multiplicative Gaussian (speckle) noise, and (\subref{fig:psnr:pepp}) salt and pepper noise (see legend for colours).}\label{fig:psnr}
\end{figure*}
\input{./tex/figure-profile}

\subsection{Response to Noise}\label{subsec:noise}
\Cref{fig:psnr} presents the performance comparison of the proposed method with the state-of-the-art approaches under the influence of three different noises: additive Gaussian, speckle and salt \& pepper.
Evidently, the proposed method has no built-in noise suppression; as expected that the effect of noise on the enhanced image is in-line with the raw image. This inherits from the noise-sensitivity in mathematical morphological and should be taken into consideration while choosing an enhancement method.

\subsection{Profile Analysis}\label{subsec:profile}

\Cref{fig:profile} illustrates bowler-hat and state-of-the-art methods responses to a simple vessel-like structure on a synthetic image. It is obvious that the value of the enhanced image at the middle of the vessel reaches a peak value and quickly drops off and decreases at the expected thickness of the vessel by the Hessian-based methods. On the other hand, the PCT-based methods are less responsive to the centreline of the vessel, while obtaining a high response to the edges due to the contrast variations. The value of the enhanced image does not significantly peak at the vessel centre, but their response does not drop off quickly since it is free from the contrast variations. 
The proposed method has both these benefits: a maximal peak value at the vessel centre-line and an enhanced response to the edges of the vessel. As a result, the reliable vessel thicknesses can be captured.

\subsection{Response to Uneven Background Illumination}\label{subsec:backillum}

\Cref{fig:illum} presents an intuitive comparison between the proposed method and other state-of-the-art methods, reg			arding the response to the uneven background illumination.
When compared with the other methods, the proposed method maintains the high responses at the junctions and seems unaffected by uneven background illuminations.

\renewcommand\wfigure{.7}
\renewcommand\bfigure{0.135}
\begin{figure}[th!]
	\centering
	\hskip-5pt
	\begin{subfigure}[t]{\bfigure\linewidth}
		\centering
		\includegraphics[width=\linewidth,keepaspectratio=true]{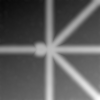}
		\caption{\quad}\label{fig:illum:orig}
	\end{subfigure}
	\begin{subfigure}[t]{\bfigure\linewidth}
		\includegraphics[width=\linewidth,keepaspectratio=true]{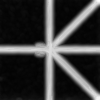}
		\caption{\quad}\label{fig:illum:proposed}
	\end{subfigure}
	\begin{subfigure}[t]{\bfigure\linewidth}
		\includegraphics[width=\linewidth,keepaspectratio=true]{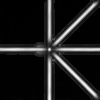}
		\caption{\quad}\label{fig:illum:vessel}
	\end{subfigure}
	\begin{subfigure}[t]{\bfigure\linewidth}
		\includegraphics[width=\linewidth,keepaspectratio=true]{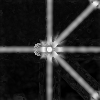}
		\caption{\quad}\label{fig:illum:neurite}
	\end{subfigure}
	\begin{subfigure}[t]{\bfigure\linewidth}
		\includegraphics[width=\linewidth,keepaspectratio=true]{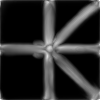}
		\caption{\quad}\label{fig:illum:pctVessel}
	\end{subfigure}
	\begin{subfigure}[t]{\bfigure\linewidth}
		\includegraphics[width=\linewidth,keepaspectratio=true]{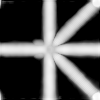}
		\caption{\quad}\label{fig:illum:pctNeurite}
	\end{subfigure}
	\begin{subfigure}[t]{\bfigure\linewidth}
		\includegraphics[width=\linewidth,keepaspectratio=true]{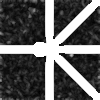}
		\caption{\quad}\label{fig:illum:vr}
	\end{subfigure}
	\caption{Comparison of the methods' abilities to deal with uneven background illumination. (\subref{fig:illum:orig}) The original image, (\subref{fig:illum:proposed}) the bowler-hat, and state-of-the-art methods respectively; 
		(\subref{fig:illum:vessel}) vesselness, 
		(\subref{fig:illum:neurite}) neuriteness,
		(\subref{fig:illum:pctVessel}) PCT vesselness,
		(\subref{fig:illum:pctNeurite}) PCT neuriteness, and
		(\subref{fig:illum:vr}) volume ratio.}\label{fig:illum}
	\vskip-15pt
\end{figure}

\input{./tex/figure-synthetic-compare2}

\subsection{Response to Vessels, Intersections/Junctions, and Blobs}\label{subsec:compare}

\Cref{fig:syn} illustrates the comparison between the proposed method and state-of-the-art methods. It is obvious that most of the state-of-the-art methods fail at the junction like in~\Cref{fig:syn:vess} and some of those create false vessels effects as in~\Cref{fig:syn:pctNeurite} or add noise the enhance image~\Cref{fig:syn:vr}. Compare to others, our proposed method is free from all of these effects and artefacts, but it is not good at suppressing the blob-like structures as like vesselness or neuriteness.\vspace{-5pt}

\makeatletter
\define@key{Gin}{mycrops}[]{\setkeys{Gin}{trim={220 120 250 120},clip}}
\makeatother
\renewcommand\bfigure{0.31}
\begin{figure}[th!]
	\centering	
	\begin{subfigure}[t]{\bfigure\linewidth}
		\includegraphics[mycrops,width=\linewidth,keepaspectratio=true]{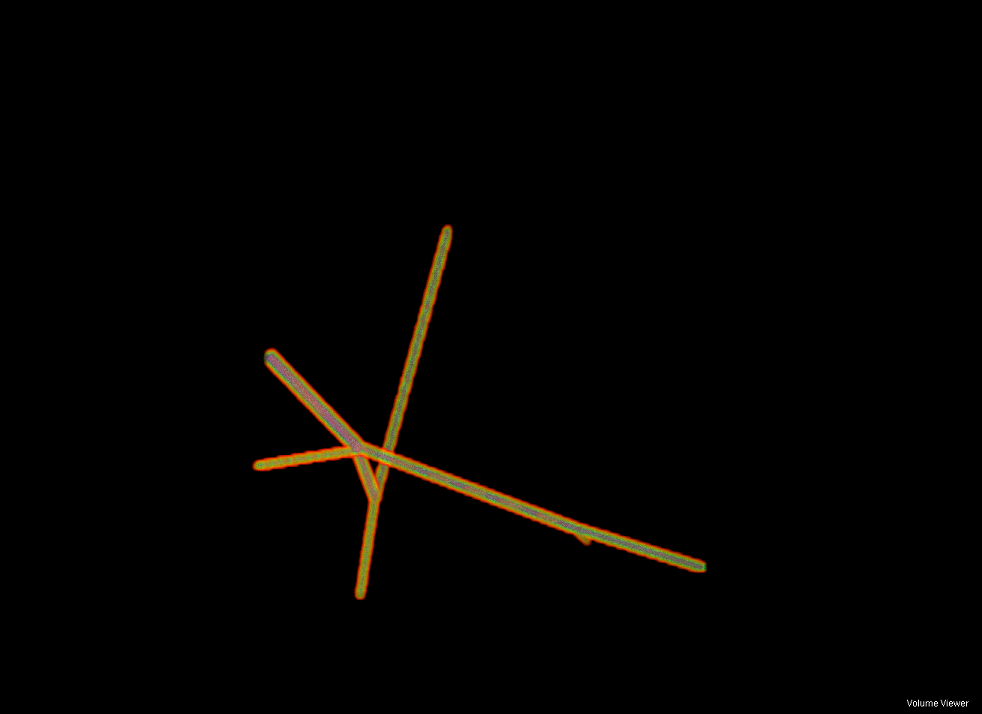}\vskip-4pt
		\caption{5}\label{fig:vascu:5}
	\end{subfigure}
	\begin{subfigure}[t]{\bfigure\linewidth}
		\includegraphics[mycrops,width=\linewidth,keepaspectratio=true]{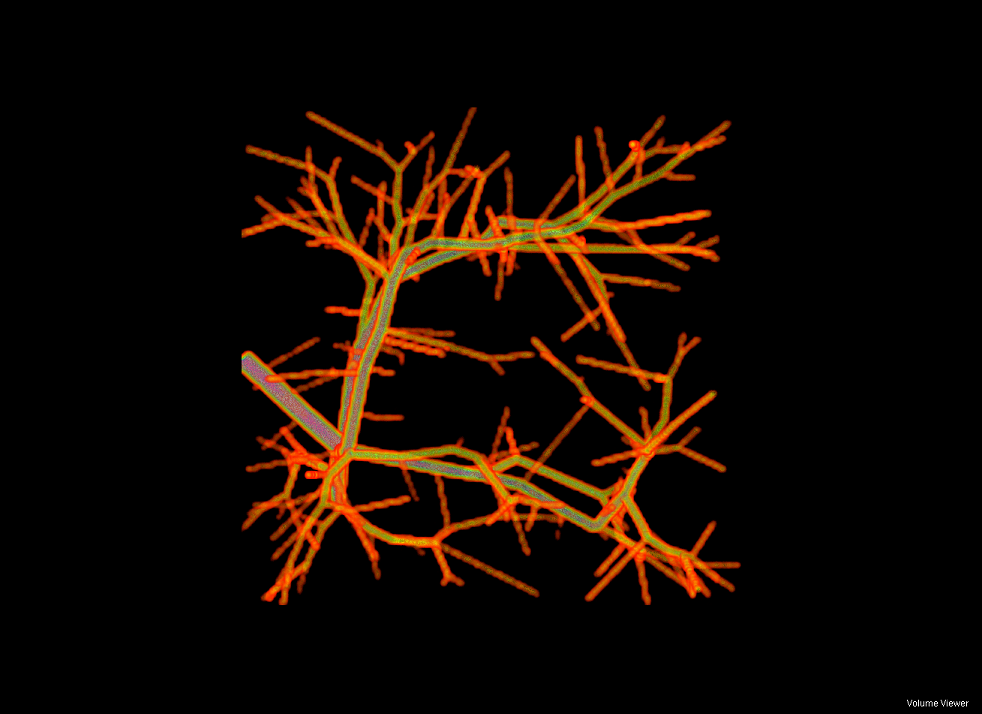}\vskip-4pt
		\caption{200}\label{fig:vascu:200}
	\end{subfigure}
	\begin{subfigure}[t]{\bfigure\linewidth}
		\includegraphics[mycrops,width=\linewidth,keepaspectratio=true]{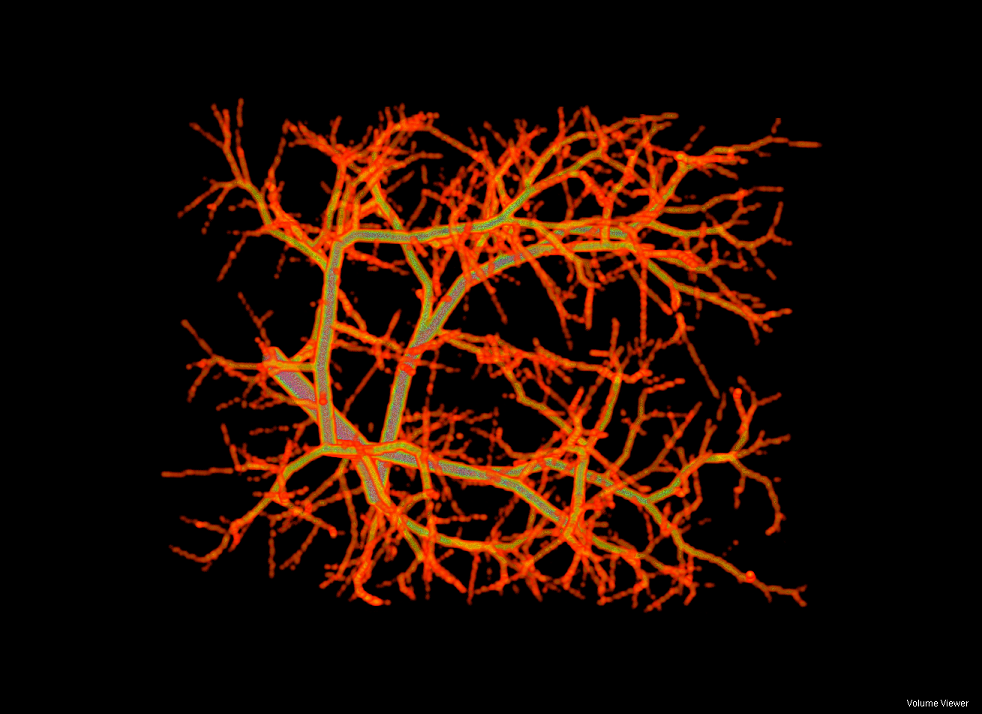}\vskip-4pt
		\caption{1000}\label{fig:vascu:1000}
	\end{subfigure}\vskip4pt
	\begin{subfigure}[t]{\bfigure\linewidth}
		\includegraphics[width=\linewidth,keepaspectratio=true]{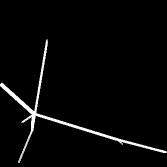}\vskip-4pt
		\caption{5}\label{fig:vascu:bow:5}
	\end{subfigure}
	\begin{subfigure}[t]{\bfigure\linewidth}
	\includegraphics[width=\linewidth,keepaspectratio=true]{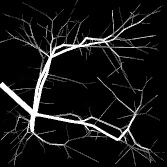}\vskip-4pt
	\caption{200}\label{fig:vascu:bow:200}
	\end{subfigure}
	\begin{subfigure}[t]{\bfigure\linewidth}
	\includegraphics[width=\linewidth,keepaspectratio=true]{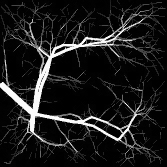}\vskip-4pt
	\caption{1000}\label{fig:vascu:bow:1000}
\end{subfigure}
	\caption{Visualisation of 3D synthetic vascular networks images generated with the VascuSynth Software~\cite{vascusynth2010}. The images~(\subref{fig:vascu:5}-\subref{fig:vascu:1000}) (167x167x167 voxels) are used to quantitatively validate the proposed method and the state-of-the-art methods and (\subref{fig:vascu:bow:5}-\subref{fig:vascu:bow:1000}) are the result of the proposed method results. More detailed results are shown in \Cref{fig:roc:vascu} and in \Cref{tab:auc}. } \label{fig:vascu:input}
\end{figure}
\begin{figure}[ht!]
	\centering
	\begin{subfigure}[t]{.6\linewidth}
		\begin{tikzpicture}
		\begin{axis}[
		width = \linewidth,
		xlabel={False Positive Rate (1-Specificity)},
		ylabel={True Positive Rate (1-Sensitivity)},
		label style={font=\footnotesize},
		tick label style={font=\footnotesize},
		ytickmin=0, ymax=1.05,
		xtickmin=0, xtickmax=1.05,
		enlargelimits=false,
		legend entries = {Raw image, Vesselness, Neuriteness, PCT vesselness, PCT neuriteness, Volume ratio, Bowler-hat},
		legend columns = 1,
		legend style={font=\footnotesize,line width=2pt, draw=none,},
		legend to name=leg3,
		grid=major, 
		grid style={dashed,gray!30}, 
		]
		\addplot[color=black, dashed,line width=1pt] table [x index=0, y index=1, col sep=comma] {image/dat-files/vascuO.dat};
		\addplot[color=brewerDark1,line width=1pt] table [x index=0, y index=1, col sep=comma] {image/dat-files/vascuV.dat};
		\addplot[color=brewerDark2,line width=1pt] table [x index=0, y index=1, col sep=comma] {image/dat-files/vascuN.dat};
		\addplot[color=brewerDark3,line width=1pt] table [x index=0, y index=1, col sep=comma] {image/dat-files/vascuPV.dat};
		\addplot[color=brewerDark4,line width=1pt] table [x index=0, y index=1, col sep=comma]{image/dat-files/vascuPN.dat};
		\addplot[color=brewerDark5,line width=1pt] table [x index=0, y index=1, col sep=comma]{image/dat-files/vascuVR.dat};
		\addplot[color=black,line width=1.5pt] table [x index=0, y index=1, col sep=comma] {image/dat-files/vascuP.dat};
		\end{axis}
		\end{tikzpicture}
	\end{subfigure}
	\begin{subfigure}{.25\linewidth}
		\vskip-100pt
	\centering
	\pgfplotslegendfromname{leg3}
\end{subfigure}
\caption{Mean ROC curve for all vascular networks images from \Cref{fig:vascu:input} calculated using the proposed and the state-of-the-art methods (see legend for colours). Individual AUC values can be found in \Cref{tab:auc}.}\label{fig:roc:vascu}
\end{figure}
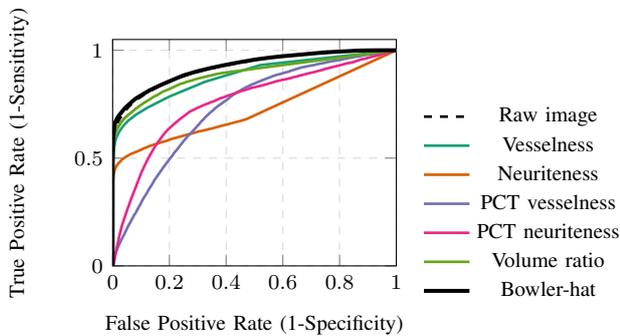

\subsection{Response to Vascular Network Complexity}
Nine volumetric images and their corresponding ground truth images of 3D synthetic vascular networks with an increasing complexity were generated using the VascuSynth Software~\cite{vascusynth2010}, as shown in~\Cref{fig:vascu:input}. 
We tested the proposed methods as well as the aforementioned other approaches on these images. The results are presented in~\Cref{tab:auc}.
\Cref{fig:roc:vascu} also demonstrates the ROC curve all over the nine enhanced images. It appears that the proposed method clearly has the highest AUC value (0.965) compare to the state-of-the-art methods. Overall, the proposed method performance is better than the state-of-the-art methods.

\begin{table}[h!]
	\centering
	\begin{adjustbox}{width=1\linewidth}
		\begin{tabular}{lcccccc}\toprule\toprule
			&\multicolumn{6}{c}{AUC}\\ \cmidrule(l){2-6}\cmidrule(l){2-7}
			{\centering Nodes}	&Vesselness &Neuriteness & PCT ves. & PCT veu.& Volume ratio & \textbf{Bowler-hat}\\
			\midrule
			5			&\textbf{0.999}	&0.923	&0.840	&0.897	&\textbf{0.999}	&\textbf{0.999}\\
			\midrule
			10			&0.996	&0.883	&0.820	&0.873	&0.998	&\textbf{0.999}\\
			\midrule
			50			&0.976	&0.830	&0.794	&0.851	&0.981	&\textbf{0.994}\\
			\midrule
			100			&0.951	&0.778	&0.778	&0.827	&0.957	&\textbf{0.982}\\
			\midrule
			200 		&0.930	&0.755	&0.770	&0.799	&0.936	&\textbf{0.966}\\
			\midrule
			400			&0.910	&0.746	&0.749	&0.788	&0.917	&\textbf{0.950}\\
			\midrule
			600			&0.902	&0.743	&0.742	&0.777	&0.909	&\textbf{0.941}\\
			\midrule
			800			&0.885	&0.719	&0.724	&0.756	&0.893	&\textbf{0.926}\\
			\midrule
			1000		&0.884	&0.722	&0.726	&0.759	&0.891	&\textbf{0.924}\\
			\midrule
			mean(std)	&0.937(0.045)	&0.788(0.073)	&0.771(0.04)	&0.814(0.05)	&0.942(0.043)	 &\textbf{0.965(0.03)}\\ \midrule
			\bottomrule
		\end{tabular}
	\end{adjustbox}
	\caption{AUC values for nine 3D image of vascular networks with increasing network's complexity (see \Cref{fig:vascu:input}) enhanced with the proposed and the state-of-the-art methods. Best results for each vascular network are in bold. ROC curve of the all volumetric images can be seen in \Cref{fig:roc:vascu}.}\label{tab:auc}
\end{table}

\renewcommand\wfigure{.7}
\renewcommand\bfigure{0.15}
\begin{figure}[h!]
	\centering
	\begin{subfigure}[t]{.015\linewidth}
		\vskip-30pt
		\textbf{1}
	\end{subfigure}
	\begin{subfigure}[t]{\bfigure\linewidth}
		\centering
		\includegraphics[width=\linewidth,keepaspectratio=true]{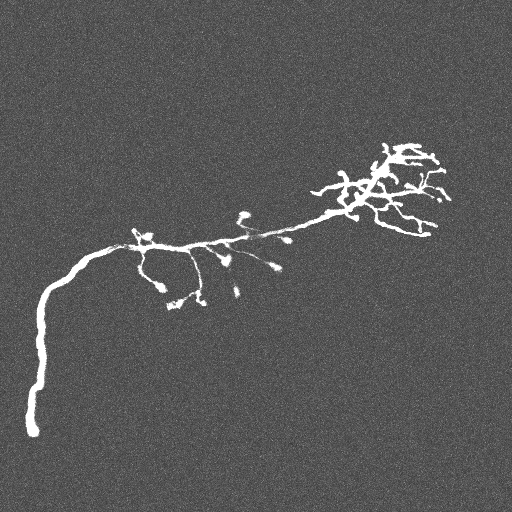}
	\end{subfigure}
	\begin{subfigure}[t]{\bfigure\linewidth}
		\includegraphics[width=\linewidth,keepaspectratio=true]{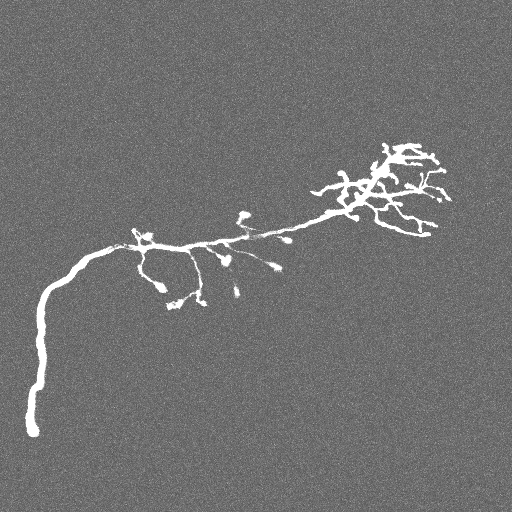}
	\end{subfigure}
	\begin{subfigure}[t]{\bfigure\linewidth}
		\includegraphics[width=\linewidth,keepaspectratio=true]{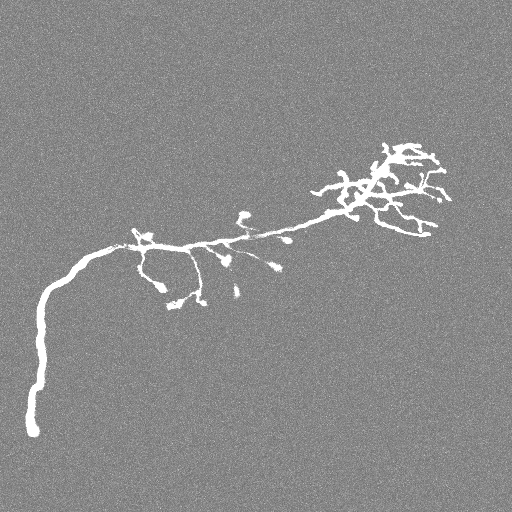}
	\end{subfigure}
	\begin{subfigure}[t]{\bfigure\linewidth}
		\includegraphics[width=\linewidth,keepaspectratio=true]{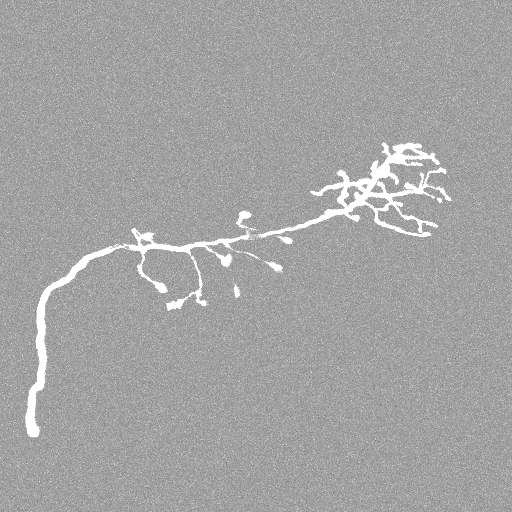}
	\end{subfigure}
	\begin{subfigure}[t]{\bfigure\linewidth}
		\includegraphics[width=\linewidth,keepaspectratio=true]{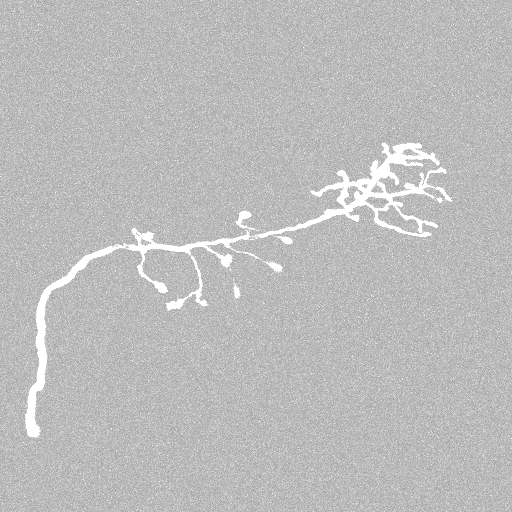}
	\end{subfigure}
	\begin{subfigure}[t]{\bfigure\linewidth}
		\includegraphics[width=\linewidth,keepaspectratio=true]{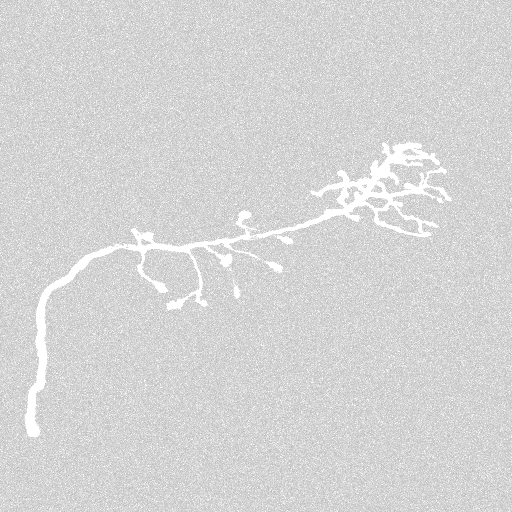}
	\end{subfigure}
\vskip4pt
	\begin{subfigure}[t]{.015\linewidth}
		\vskip-30pt
		\textbf{2}
	\end{subfigure}
	\begin{subfigure}[t]{\bfigure\linewidth}
		\centering
		\includegraphics[width=\linewidth,keepaspectratio=true]{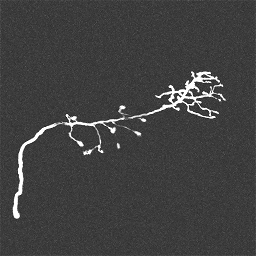}\vskip-4pt
		\caption*{AUC:0.959}
	\end{subfigure}
	\begin{subfigure}[t]{\bfigure\linewidth}
		\includegraphics[width=\linewidth,keepaspectratio=true]{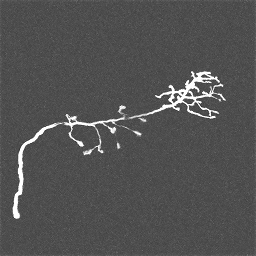}\vskip-4pt
		\caption*{AUC:0.959}
	\end{subfigure}
	\begin{subfigure}[t]{\bfigure\linewidth}
		\includegraphics[width=\linewidth,keepaspectratio=true]{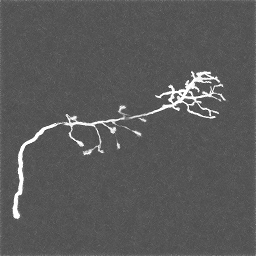}\vskip-4pt
		\caption*{AUC:0.955}
	\end{subfigure}
	\begin{subfigure}[t]{\bfigure\linewidth}
		\includegraphics[width=\linewidth,keepaspectratio=true]{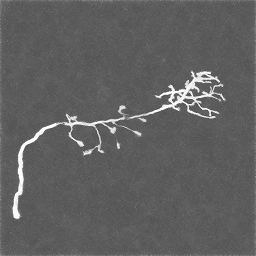}\vskip-4pt
		\caption*{AUC:0.953}
	\end{subfigure}
	\begin{subfigure}[t]{\bfigure\linewidth}
		\includegraphics[width=\linewidth,keepaspectratio=true]{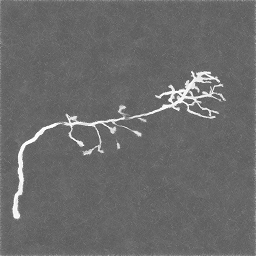}\vskip-4pt
		\caption*{AUC:0.951}
	\end{subfigure}
	\begin{subfigure}[t]{\bfigure\linewidth}
		\includegraphics[width=\linewidth,keepaspectratio=true]{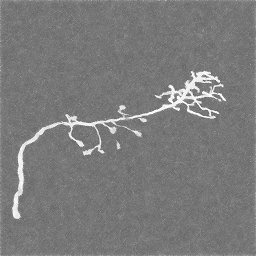}\vskip-4pt
		\caption*{AUC:0.951}
	\end{subfigure}
	\begin{subfigure}[t]{.015\linewidth}
		\vskip-30pt
		\textbf{3}
	\end{subfigure}
	\begin{subfigure}[t]{\bfigure\linewidth}
		\centering
		\includegraphics[width=\linewidth,keepaspectratio=true]{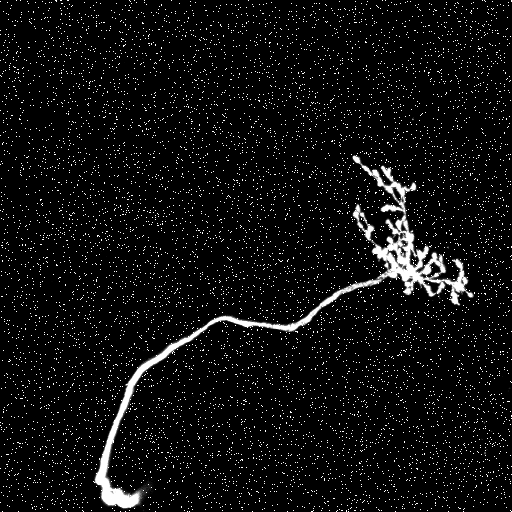}
	\end{subfigure}
	\begin{subfigure}[t]{\bfigure\linewidth}
		\includegraphics[width=\linewidth,keepaspectratio=true]{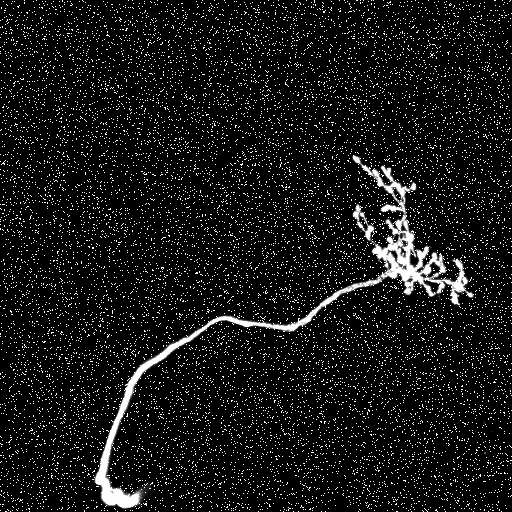}
	\end{subfigure}
	\begin{subfigure}[t]{\bfigure\linewidth}
		\includegraphics[width=\linewidth,keepaspectratio=true]{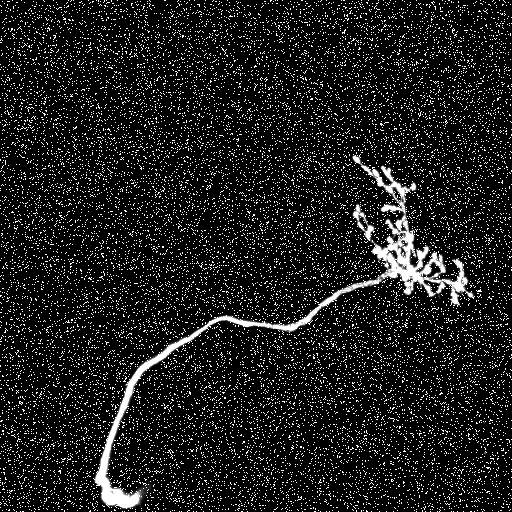}
	\end{subfigure}
	\begin{subfigure}[t]{\bfigure\linewidth}
		\includegraphics[width=\linewidth,keepaspectratio=true]{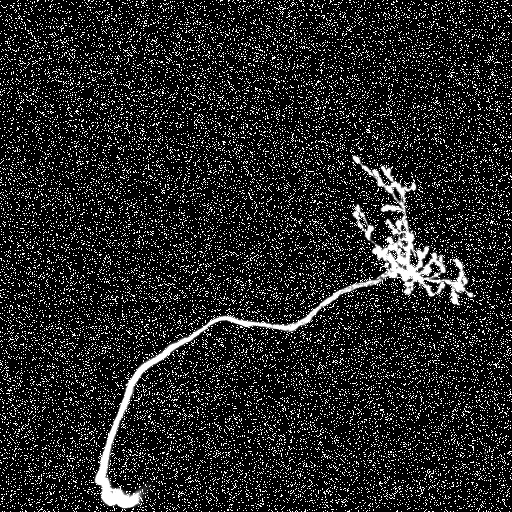}
	\end{subfigure}
	\begin{subfigure}[t]{\bfigure\linewidth}
		\includegraphics[width=\linewidth,keepaspectratio=true]{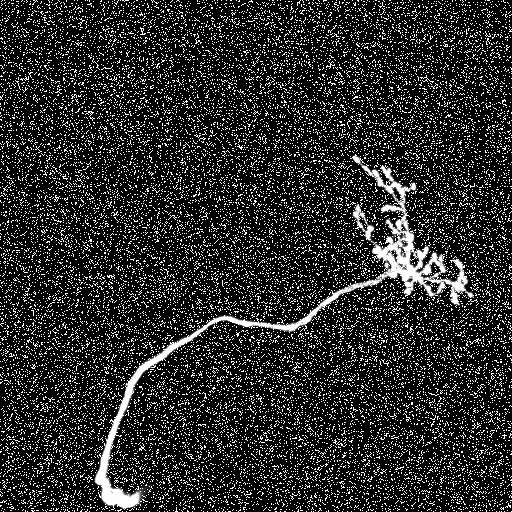}
	\end{subfigure}
	\begin{subfigure}[t]{\bfigure\linewidth}
		\includegraphics[width=\linewidth,keepaspectratio=true]{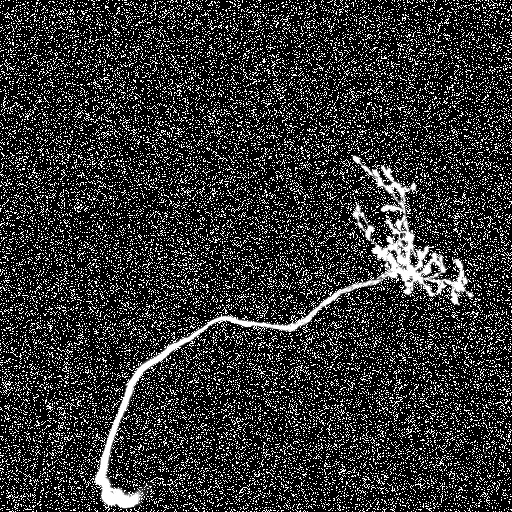}
	\end{subfigure}
\vskip5pt
	\begin{subfigure}[t]{.015\linewidth}
		\vskip-30pt
		\textbf{4}
	\end{subfigure}
	\begin{subfigure}[t]{\bfigure\linewidth}
		\centering
		\includegraphics[width=\linewidth,keepaspectratio=true]{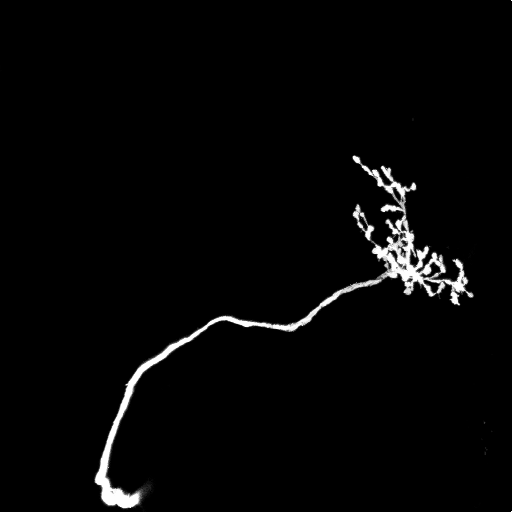}\vskip-4pt
		\caption*{AUC:0.953}
		\caption{lvl = 10}\label{fig:fiber:sigma10}
	\end{subfigure}
	\begin{subfigure}[t]{\bfigure\linewidth}
		\includegraphics[width=\linewidth,keepaspectratio=true]{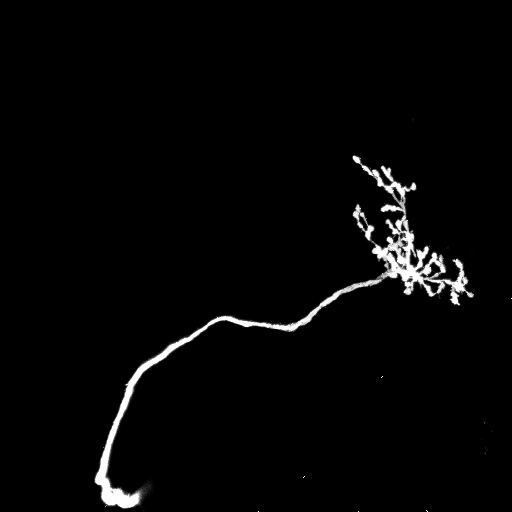}\vskip-4pt
		\caption*{AUC:0.953}
		\caption{lvl = 20}\label{fig:fiber:sigma20}
	\end{subfigure}
	\begin{subfigure}[t]{\bfigure\linewidth}
		\includegraphics[width=\linewidth,keepaspectratio=true]{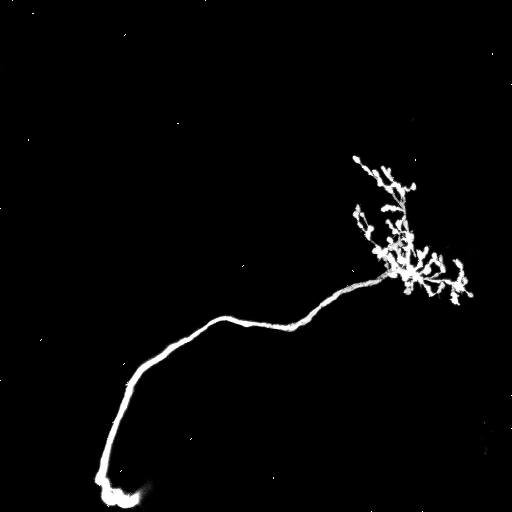}\vskip-4pt
		\caption*{AUC:0.953}
		\caption{lvl = 30}\label{fig:fiber:sigma30}
	\end{subfigure}
	\begin{subfigure}[t]{\bfigure\linewidth}
		\includegraphics[width=\linewidth,keepaspectratio=true]{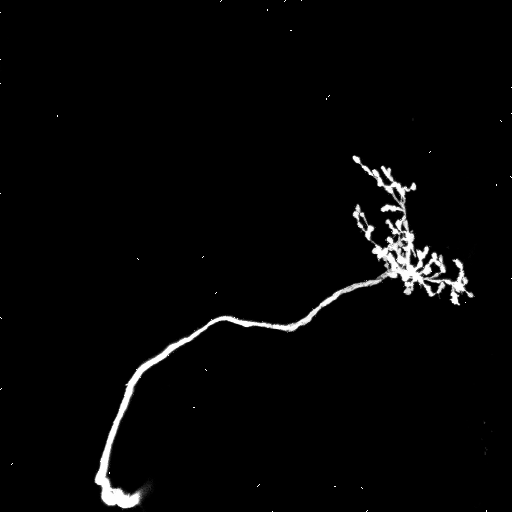}\vskip-4pt
		\caption*{AUC:0.953}
		\caption{lvl = 40}\label{fig:fiber:sigma40}
	\end{subfigure}
	\begin{subfigure}[t]{\bfigure\linewidth}
		\includegraphics[width=\linewidth,keepaspectratio=true]{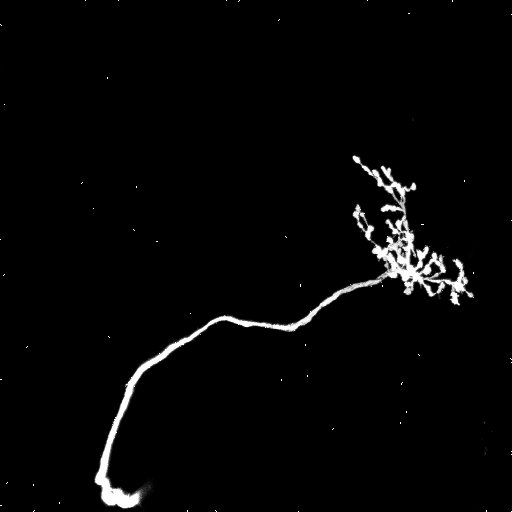}\vskip-4pt
		\caption*{AUC:0.953}
		\caption{lvl = 50}\label{fig:fiber:sigma50}
	\end{subfigure}
	\begin{subfigure}[t]{\bfigure\linewidth}
		\includegraphics[width=\linewidth,keepaspectratio=true]{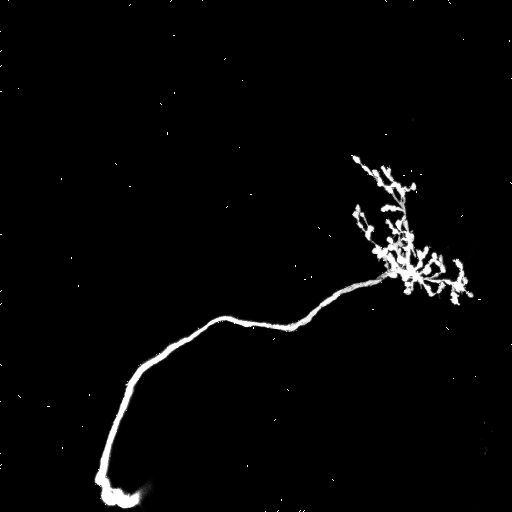}\vskip-4pt
		\caption*{AUC:0.953}
		\caption{lvl = 60}\label{fig:fiber:sigma60}
	\end{subfigure}
	\caption{Application of the proposed method into the Olfactory Projection Neuron dataset from the DIADEM Challenge. All of the images are 2D maximum intensity projections. 1,3) Input images that have been contaminated by different levels (increasing left to right) and types of noise (1 - Guassian additive noise; 3 - salt and pepper noise). 2,4) Enhancement results with the proposed method and corresponding AUC values.}\label{fig:fiber}
	\vskip-10pt
\end{figure}

\subsection{Real Data}\label{subsec:real}

An Olfactory Projection Fibers image dataset from DIADEM Challenge~\cite{brown2011diadem} is used to demonstrate the robustness of proposed method against the noise. In two exemplary fibers images, a Gaussian noise was introduced at the noise levels ranging from $\sigma=10$ to $\sigma=60$, and  salt and pepper noise at the different level of density  $\rho=10$ to $\rho=60$ see \Cref{fig:fiber}. Such images were then enhanced with the proposed method and the AUC values were calculated and presented in \Cref{fig:fiber}.  
We also tested the performance of the proposed method on 3D real images. Here we adopt three representative types of real images, namely microcomputers network in plant cell, keratin network in skin cell, and neuronal network. Correspondingly we compare the output of the proposed method with five other approaches, and the results are shown in~\Cref{fig:proposedResult}. It is clearly suggested that our method has the best performance in preserving  junctions.

\renewcommand\hfigure{.97}
\renewcommand\wfigure{.95}
\renewcommand\bfigure{0.15}
\renewcommand\sfigure{0.24}
\begin{figure} [t!]
	\centering
	\begin{subfigure}[t]{\sfigure\linewidth}
		\begin{tikzpicture}
		\node[anchor=south west,inner sep=0] (image) at (0,0) 
		{\includegraphics[trim={0px 0px 0px 30px}, clip,width=\wfigure\linewidth,keepaspectratio=true]{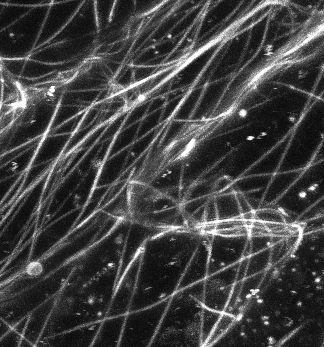}};
		\begin{scope}[x={(image.south east)},y={(image.north west)}]
		\draw[red,thick,] (0.8,0.9) rectangle (0.5,0.6);
		\end{scope}
		\end{tikzpicture}\vskip-4pt
		\caption{\quad}\label{fig:OrCel}
	\end{subfigure}
		\begin{subfigure}[t]{\sfigure\linewidth}
		\begin{tikzpicture}
		\node[anchor=south west,inner sep=0] (image) at (0,0) 
		{\includegraphics[width=\wfigure\linewidth,keepaspectratio=true]{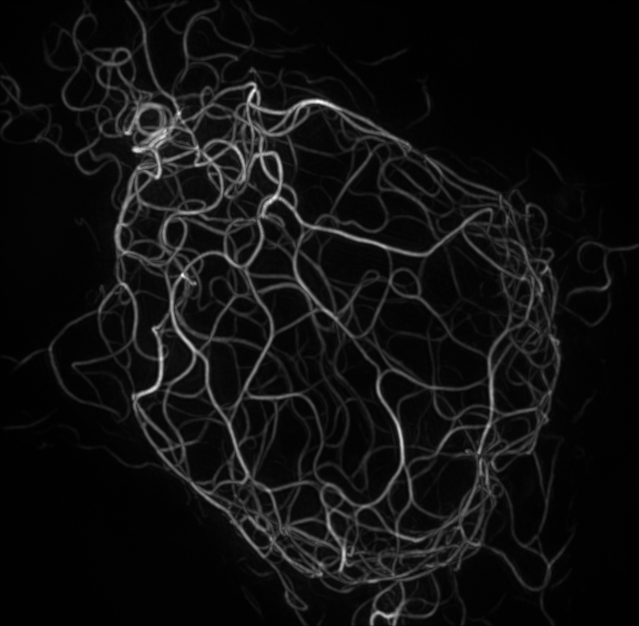}};
		\begin{scope}[x={(image.south east)},y={(image.north west)}]
		\draw[red,thick,] (0.48,0.3) rectangle (0.83,0.67);
		\end{scope}
		\end{tikzpicture}\vskip-4pt
		\caption{\quad}\label{fig:OrKer}	
	\end{subfigure}
		\begin{subfigure}[t]{\sfigure\linewidth}
		\begin{tikzpicture}
		\node[anchor=south west,inner sep=0] (image) at (0,0) 
		{\includegraphics[trim={10px 100px 10px 0px}, clip, width=\wfigure\linewidth, keepaspectratio=true]{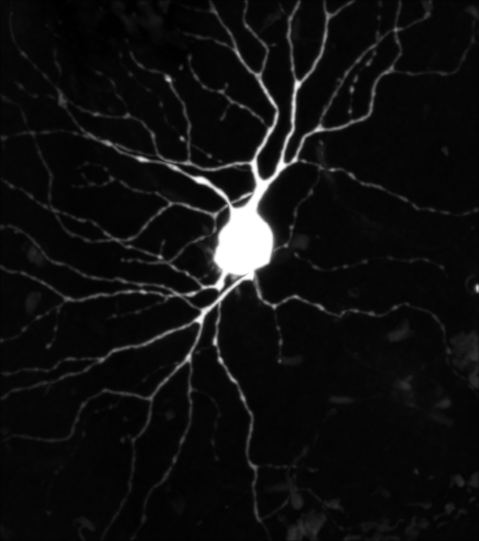}};
		\begin{scope}[x={(image.south east)},y={(image.north west)}]
		\draw[red,thick,] (0.2,0.38) rectangle (0.8,0.95);
		\end{scope}
		\end{tikzpicture}\vskip-4pt
		\caption{\quad}\label{fig:OrNeurite}
	\end{subfigure}\\
	\begin{subfigure}[t]{\bfigure\linewidth}
		\includegraphics[clip,width=\wfigure\linewidth,keepaspectratio=true]{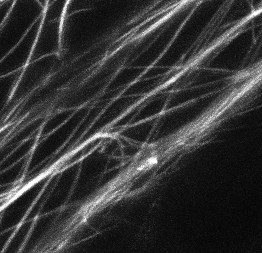}
	\end{subfigure}
	\begin{subfigure}[t]{\bfigure\linewidth}
		\includegraphics[trim={300px 200px 100px 200px},clip,width=\hfigure\linewidth,keepaspectratio=true]{keratin2}\vskip-4pt
		\caption{ROIs}\label{fig:proposedResult:Or}
	\end{subfigure}
	\begin{subfigure}[t]{\bfigure\linewidth}
		\includegraphics[trim={100px 270px 100px 0px},clip,width=\wfigure\linewidth,keepaspectratio=true]{LuciferS.png}\vskip-4pt
	\end{subfigure}\\
	\begin{subfigure}[t]{\bfigure\linewidth}
	\includegraphics[trim={0px 0px 00px 00px}, 
	clip,width=\wfigure\linewidth,keepaspectratio=true]{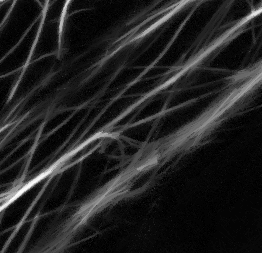}\vskip-4pt
	\end{subfigure}
	\begin{subfigure}[t]{\bfigure\linewidth}
	\includegraphics[trim={300px 200px 100px 200px}, 
	clip,width=\hfigure\linewidth,keepaspectratio=true]{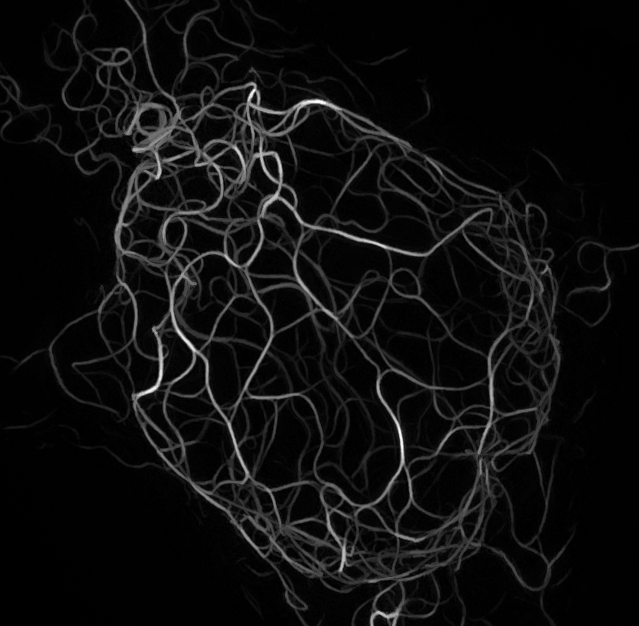}\vskip-4pt
		\caption{\quad}\label{fig:proposedResult:Granulo}
	\end{subfigure}
	\begin{subfigure}[t]{\bfigure\linewidth}
		\includegraphics[trim={100px 270px 100px 0px}, 
		clip,width=\wfigure\linewidth,keepaspectratio=true]{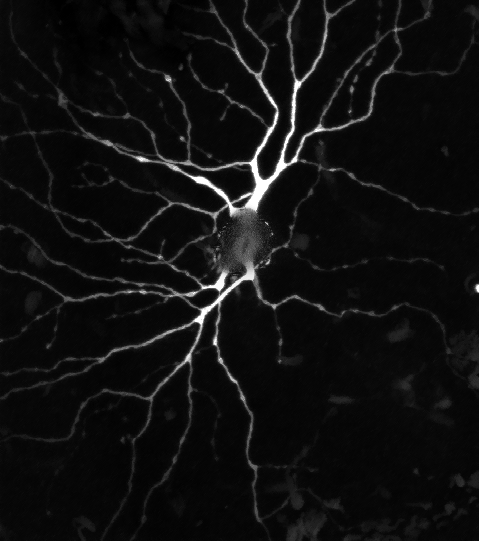}\vskip-4pt
	\end{subfigure}
\hspace{1mm}
	\begin{subfigure}[t]{\bfigure\linewidth}
		\includegraphics[clip,width=\wfigure\linewidth,keepaspectratio=true]{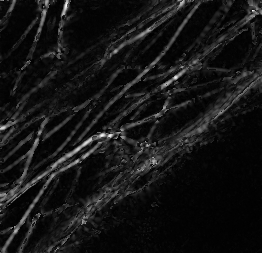}\vskip-4pt
	\end{subfigure}
	\begin{subfigure}[t]{\bfigure\linewidth}
		\includegraphics[trim={300px 200px 100px 200px},clip,width=\hfigure\linewidth,keepaspectratio=true]{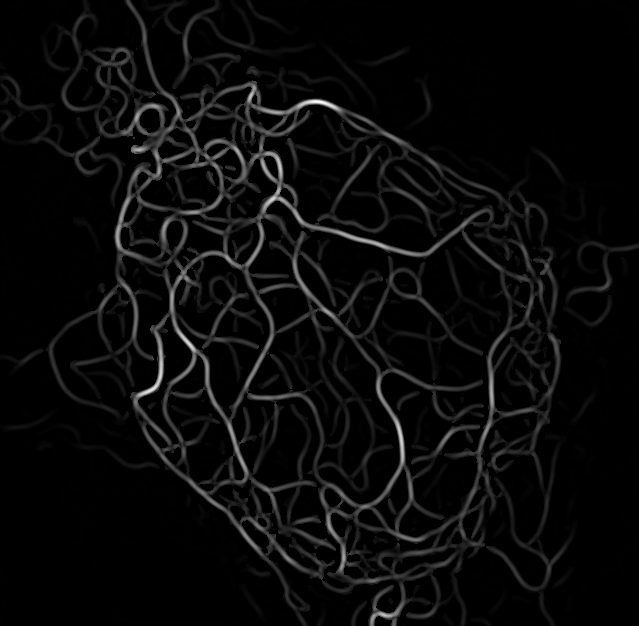}\vskip-4pt
		\caption{\quad}\label{fig:proposedResult:Neuriteness}
	\end{subfigure}
	\begin{subfigure}[t]{\bfigure\linewidth}
		\includegraphics[trim={100px 270px 100px 0px} ,clip,width=\wfigure\linewidth,keepaspectratio=true]{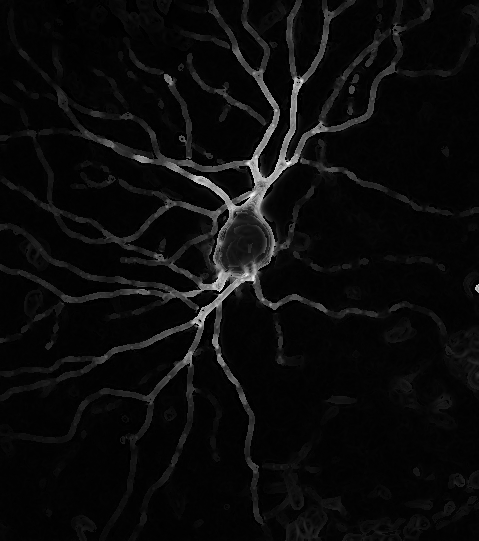}\vskip-4pt
	\end{subfigure}
	\begin{subfigure}[t]{\bfigure\linewidth}
		\includegraphics[trim={0px 0px 0px 0px},clip,width=\wfigure\linewidth,keepaspectratio=true]{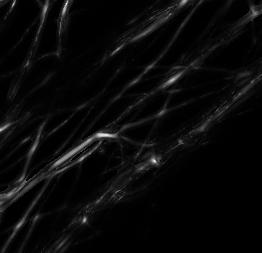}
	\end{subfigure}
	\begin{subfigure}[t]{\bfigure\linewidth}
		\includegraphics[trim={300px 200px 100px 200px},clip,width=\hfigure\linewidth,keepaspectratio=true]{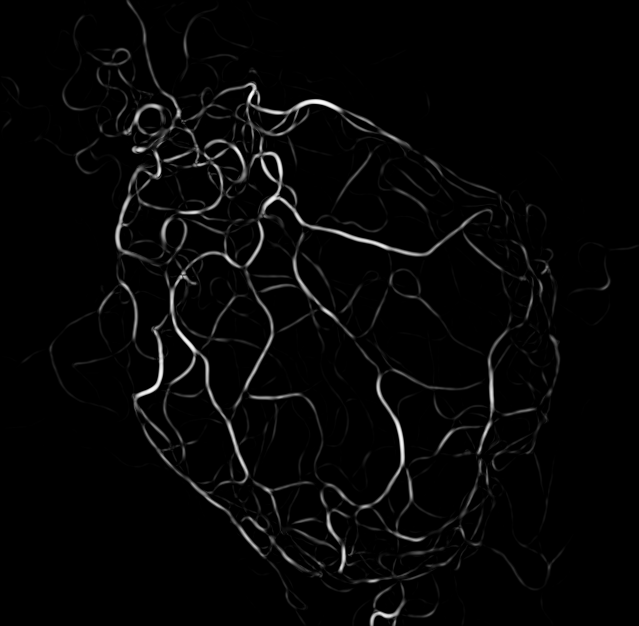}\vskip-4pt
		\caption{\quad}\label{fig:proposedResult:Vesselness}
	\end{subfigure}
	\begin{subfigure}[t]{\bfigure\linewidth}
		\includegraphics[trim={100px 270px 100px 0px}, clip,width=\wfigure\linewidth,keepaspectratio=true]{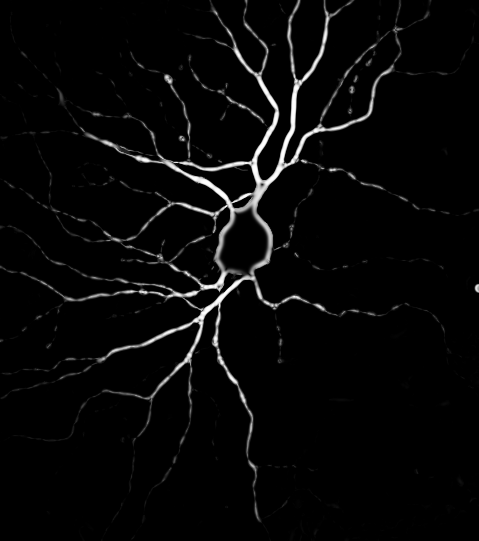}\vskip-4pt
	\end{subfigure}
\hspace{1mm}
	\begin{subfigure}[t]{\bfigure\linewidth}
		\includegraphics[trim={0px 0px 0px 0px}, 
		clip,width=\wfigure\linewidth,keepaspectratio=true]{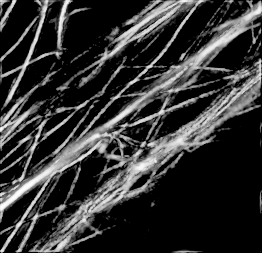}\vskip-4pt
	\end{subfigure}
	\begin{subfigure}[t]{\bfigure\linewidth}
		\includegraphics[trim={300px 200px 100px 200px}, 
		clip,width=\hfigure\linewidth,keepaspectratio=true]{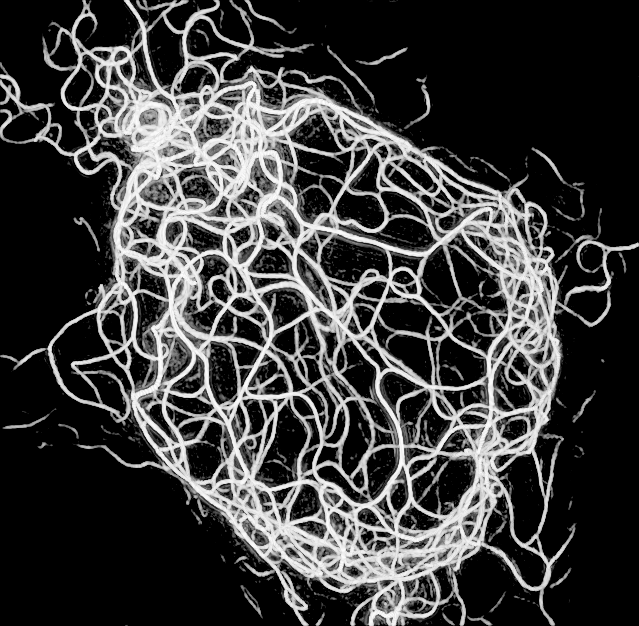}\vskip-4pt
		\caption{\quad}\label{fig:proposedResult:PCTNe}
	\end{subfigure}
	\begin{subfigure}[t]{\bfigure\linewidth}
		\includegraphics[trim={100px 270px 100px 0px}, 
		clip,width=\wfigure\linewidth,keepaspectratio=true]{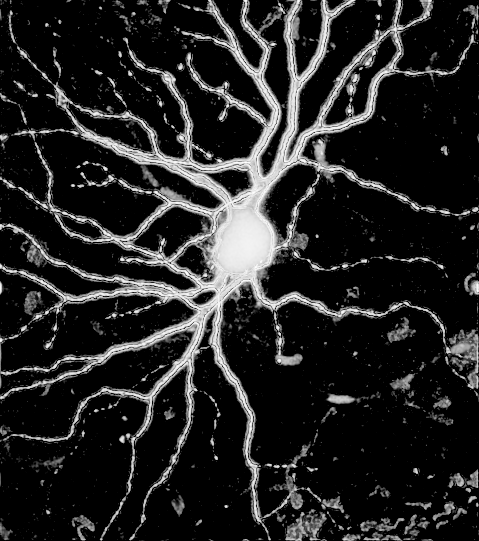}\vskip-4pt
	\end{subfigure}
	\begin{subfigure}[t]{\bfigure\linewidth}
		\includegraphics[trim={0px 0px 00px 00px}, 
		clip,width=\wfigure\linewidth,keepaspectratio=true]{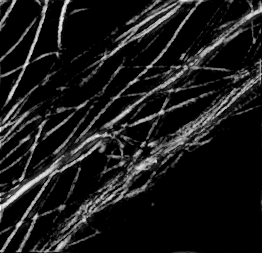}\vskip-4pt
	\end{subfigure}
	\begin{subfigure}[t]{\bfigure\linewidth}
		\includegraphics[trim={300px 200px 100px 200px}, 
		clip,width=\hfigure\linewidth,keepaspectratio=true]{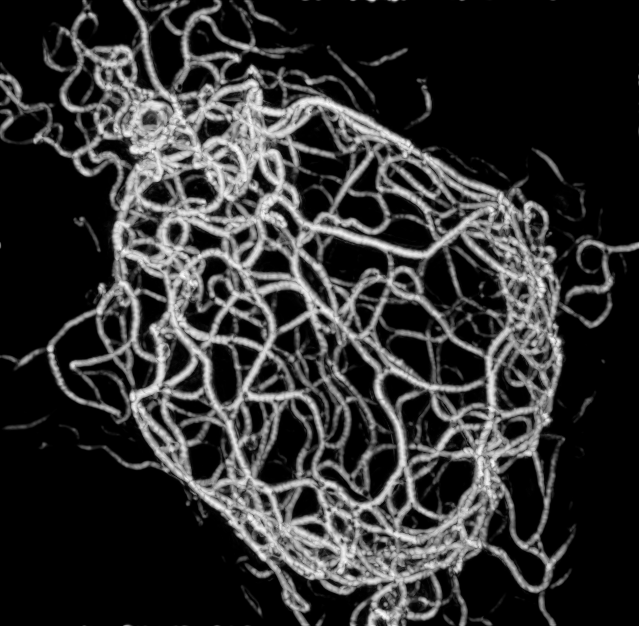}\vskip-4pt
		\caption{\quad}\label{fig:proposedResult:PCTVes}
	\end{subfigure}
	\begin{subfigure}[t]{\bfigure\linewidth}
		\includegraphics[trim={100px 270px 100px 0px}, 
		clip,width=\wfigure\linewidth,keepaspectratio=true]{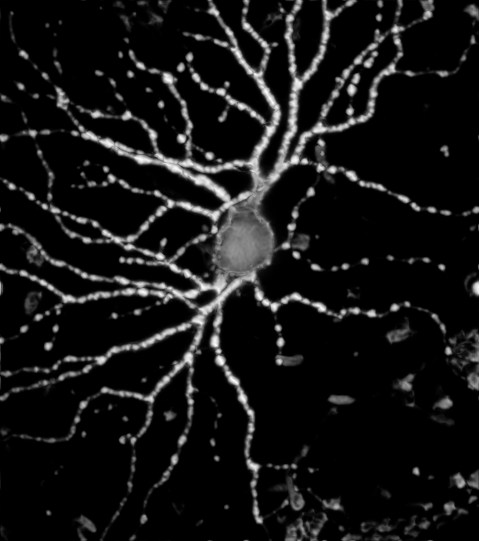}\vskip-4pt
	\end{subfigure}
\hspace{1mm}
	\begin{subfigure}[t]{\bfigure\linewidth}
		\includegraphics[trim={0px 0px 00px 00px}, 
		clip,width=\wfigure\linewidth,keepaspectratio=true]{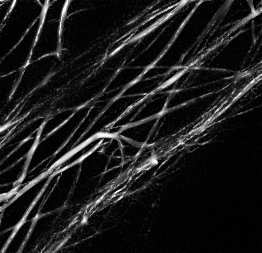}\vskip-4pt
	\end{subfigure}
	\begin{subfigure}[t]{\bfigure\linewidth}
		\includegraphics[trim={300px 200px 100px 200px},
		clip,width=\hfigure\linewidth,keepaspectratio=true]{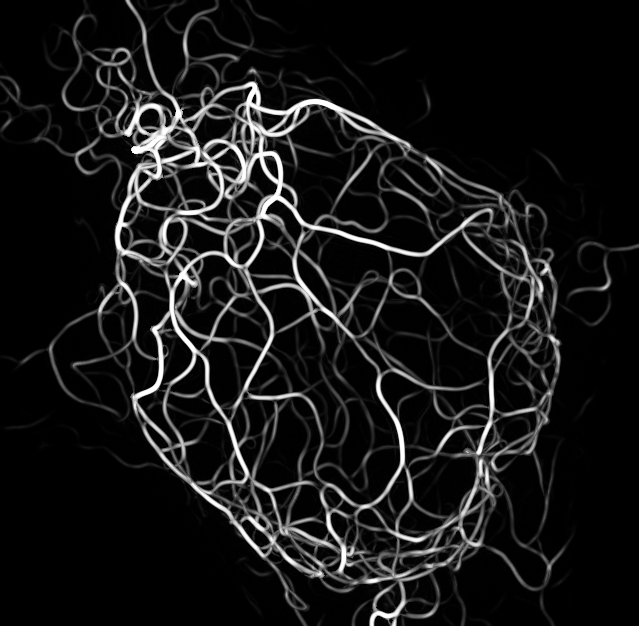}\vskip-4pt
		\caption{\quad}\label{fig:proposedResult:VR}
	\end{subfigure}
	\begin{subfigure}[t]{\bfigure\linewidth}
	\includegraphics[trim={100px 270px 100px 0px}, 
	clip,width=\wfigure\linewidth,keepaspectratio=true]{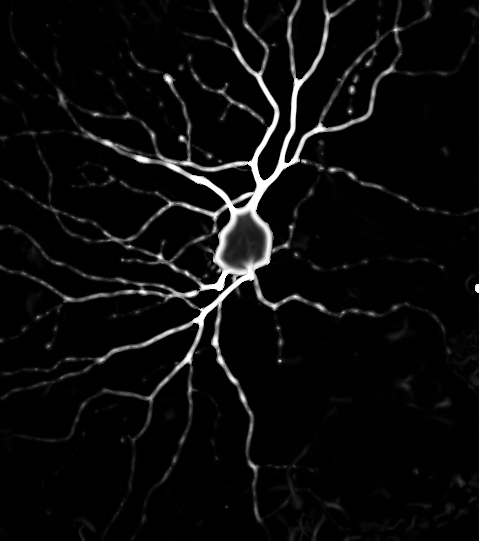}\vskip-4pt
	\end{subfigure}
	\caption {Comparison of the proposed and the state-of-the-art methods on a set of real biomedical images. 2D max projections of 3D images of microtubules network in plant cell (\subref{fig:OrCel}), keratin network in skin cell (\subref{fig:OrKer}) (provided by Dr Tim Hawkins, Durham University, UK),  and neuronal network (\subref{fig:OrNeurite}) (provided by Dr Chris Banna, UC Santa Barbara, USA). Regions of interest are highlighted in red and presented in (\subref{fig:proposedResult:Or}). Results: 
	bowler-hat (\subref{fig:proposedResult:Granulo}), neuriteness (\subref{fig:proposedResult:Neuriteness}), vesselness (\subref{fig:proposedResult:Vesselness}), PCT neuriteness (\subref{fig:proposedResult:PCTNe}), PCT vesselness (\subref{fig:proposedResult:PCTVes}) and volume ratio (\subref{fig:proposedResult:VR}).}
	\label{fig:proposedResult}\vspace{-10pt}
\end{figure}
\section{Conclusion }\label{sec:conclusion}
Hessian- or Phase Congruency Tensor-based image enhancement methods had been commonly used to enhance vessel-like structures in 3D biomedical images using measurements like vesselness, neuriteness and volume ratio. 

This paper proposes a novel mathematical morphology-based method for vessel-like structures enhancement in 3D biomedical images. The proposed method is shown to have benefits over existing methods, including no loss of signal and junctions and minimized artifacts at vessel ends. We show efficiency on both synthetic and real image datasets.

Future continuations of this work will introduce the implementation of the blob-enhancing variants of this concept.

\bibliographystyle{IEEEtran}
\bibliography{bib/reference}
\end{document}